\pdfoutput=1

\documentclass{jdmdh}
\usepackage{array}
\usepackage{pgfplots}
\pgfplotsset{compat=1.18}

\titlespacing*{\section}
{0pt}{2.5ex plus 1ex minus .2ex}{1.3ex plus .2ex}
\titlespacing*{\subsection}
{0pt}{1.0ex plus 1ex minus .2ex}{1.3ex plus .2ex}
\titlespacing*{\subsubsection}
{0pt}{1.0ex plus 1ex minus .2ex}{1.3ex plus .2ex}

\title{Affect as a proxy for literary mood}
\author[1]{Emily {\"O}hman}
\author[2]{Riikka Rossi}
\affil[1]{Waseda University, Japan} 
\affil[2]{University of Helsinki, Finland} 

\corrauthor{Emily {\"O}hman}{ohman@waseda.jp}

\usepackage{tablefootnote}
\usepackage{latexsym}
\usepackage{adjustbox}
\usepackage[T1]{fontenc}

\usepackage[utf8]{inputenc}

\usepackage{microtype}

\begin{document}
\maketitle
\begin{abstract}

We propose to use affect as a proxy for mood in literary texts. In this study, we explore the differences in computationally detecting tone versus detecting mood. Methodologically we utilize affective word embeddings to look at the affective distribution in different text segments. We also present a simple yet efficient and effective method of enhancing emotion lexicons to take both semantic shift and the domain of the text into account producing real-world congruent results closely matching both contemporary and modern qualitative analyses.
\end{abstract}

\section{Introduction}

In this study, we explore how the literary concept of mood can be studied and detected with computational methods. We propose to use affect as a proxy for mood and test our hypothesis first quantitatively on different segmentations of the different texts, and then qualitatively against expert close-readings of the same texts. We use multiple different natural language processing (NLP) approaches to attempt to identify the origins, construction, and location of mood. For this purpose, we have collected a corpus of nearly 1000 literary works published in Finnish around the year 1900. Our paper utilizes many common NLP methods and approaches from computational literary studies and affective computing, but besides our pilot study \citep{ohman2022computational}, to our knowledge, it is the first study to combine emotion detection/sentiment analysis with the study of mood in texts.

We focus on \textit{mood} as it is one of the more ephemeral yet pervasive aspects of a literary text. Mood is typically described as the atmosphere that the author creates through their word choices, style, and use of imagery and can sometimes even include \textit{tone}. The line between tone and mood can be difficult to draw, particularly when approaching the topic with computational tools, but succinctly the difference can be explained as \textit{mood} being about how the reader feels about the text, and \textit{tone} about how the implied author feels about it and uses words to convey their own attitude towards a topic or subject\footnote{On the concepts of tone and mood in literary studies, see \citet{richards,ngai2005ugly,flatley}.} \citep{turco2020book}.

Although many different literary tools contribute to the creation of mood, perhaps the most important one, that influences all the other tools as well, is the choice of words. Typically in literature, the intentionality of word choice is higher than in most other text genres (e.g. social media posts) \citep{mitchell2008intention,ohman2020xed}. It is therefore a great subject for analyzing the use and intensity of emotion-associated words in text. As there are so many components that help create mood, in a larger sense mood is not reducible to a single aspect, but generated by a set of textual elements, however, we can use the importance of word choice to our advantage. If we look at the affective distribution of the words in texts using different text segments, we can attempt to pinpoint where in a text mood is created and how that links to affect.

We suggest that the computational study of the valence of the lexicon of a literary text can be valuable in providing an accurate picture of the distribution of the positive and negative valence in a text continuum, and thus help us to better understand the relationship between the linguistic qualities of a text and its perceived emotional effects, particularly the mood of a text. Furthermore, we contribute to the discussion of the most suitable tools for interdisciplinary work including the debate about lexicon-based methods versus machine learning as the most "accurate" \citep{van2021validity,ohman2021lexval,teodorescu2022frustratingly}.

\section{Background \& Previous Work}
Around the turn of the millennium the "affective turn" took place \citep{smith2011postmodernism,armstrong2014affective}. It was a shift in attitude regarding the importance of affect in humanities and social science research, including literary studies \citep{kim2007affective}. Literature can be considered a domain where the affective functions of language are of principal importance \citep{hogan2011literature}. The affective turn has led to a significant increase in research that focuses on the affective side of text in literature \citep{armstrong2014affective}. 

The affective power of literary texts has been acknowledged since Aristotle's \textit{Poetics}, but in the 20th century, many schools of thought such as formalism, new criticism, structuralism and post-structuralism orientated the attention to formal and structural aspects of texts, whereas the study of emotions was excluded and considered as susceptible to researchers' subjective emotions. 
Research topics range from the study of literature and empathy \citep{keen2007empathy} to the study of literature and cognition \citep{hogan2011literature}), negative affects and tone in texts \citep{ngai2005ugly} to empirical perspectives \citep{54759120130101,van2018difficult}, and even emotions specific to Finnish literature \citep{rossi2020,rossi2022pohjoisia}. 

Recent studies on emotions in Finnish literature demonstrate that Finnish literature presents us with a rich body of work for developing the general theory of literature and emotions and to study the ways in which genre-specific emotional effects vary culturally and historically \citep{rossi2022pohjoisia}. While this research has opened up new perspectives it has also demonstrated there are a number of gaps and complex questions to resolve. Along with the affective turn, the question of a text's overall emotional tone or mood has aroused vivid interest (e.g. Ngai \citealt{ngai2005ugly}, \citealt{lyyti2017written}, and Rossi \citealt{rossi2020}). However, a systematic theory of how tone and mood are created and triggered in the reader is still in the works.

We suggest that a study of the emotional valence of the lexicon measured quantitatively provides a new approach that can help with understanding the components of a text's mood \citep{ohman2020litemo,ohman2022computational}. Furthermore, it has been shown that lexicon-based methods can achieve better results in emotion classification tasks than machine learning models, especially when the text segment size is optimized \citep{ohman2021lexval,teodorescu2022frustratingly}.

Parallel to the affective turn, sentiment analysis became an active field of research within natural language processing and computer science \citep{MANTYLA201816}. Sentiment analysis and emotion detection has been used with literary works with varying levels of success; \citet{kim2018survey} provide a substantive overview of sentiment analysis and emotion detection as it is used in CLS. Although many of the papers cited in the review are interesting and innovative, virtually none of them deal with topics that are common in more traditional literary studies. Common CLS topics are genre classification by emotion, story-type, sentiment tracking, and sentiment recognition (see e.g. \citealt{sprugnoli2016towards}; \citealt{schmidt2018evaluation}; \citealt{amano2023analysis}). 

As interesting and innovative as previous CLS work is, it is typically of little use to literary scholars studying affect. These CLS approaches rarely work in harmony with literary analysis in the traditional sense and usually do not even touch upon the topics that interest literary scholars such as tone, mood, and emotion evocation. We hope that our efforts will contribute to merging the talents and knowledge within CLS, digital humanities, NLP, as well as traditional literary studies.

Although there are exceptions (see e.g. \citealt{hu2021dynamic} and \citealt{herrmann2019linguistic}), it is somewhat rare for studies within the field of CLS to have literary experts working on the project, and many such projects rely heavily on the analysis of the quantitative results conducted by experts of NLP rather than experts of literature. This is not a problem in only CLS, but in many other interdisciplinary fields, particularly those with a computational element \citep{bartlett2018locus}. This is why we think it is imperative to conduct CLS (and other interdisciplinary studies) with domain experts and not just NLP knowledge and literary data.

CLS as a field has been criticized for providing either obvious results or ephemeral results that are not robust enough for repeat analysis \citep{da2019computational}. Furthermore, there seems to be a pervasive belief that state-of-the-art methods from NLP are the most accurate when used on non-standard unstructured data regardless of the research question of the downstream application of such methods. It is commonly suggested that machine learning methods are more accurate than lexicon-based ones (see e.g. \citealt{van2021validity}), but several recent papers have suggested this is not the case, especially when dealing with emotional arcs and ideal bin sizes in verbose domains \citep{ohman2021lexval,teodorescu2022frustratingly}. We have taken the criticism of both camps to heart in an effort to produce robust and reliable but also useful and interpretable results. 

\section{Data}
Our data collection was simple and straightforward. Although there are R packages and Python libraries in existence for handling Project Gutenberg downloads, Project Gutenberg discourages mass downloads using such methods, thus we used their recommended method for filtering works using http queries and downloading a set smaller amount of books at a time. We downloaded the first\footnote{Presumably in order of entry to the database.} 1000 books from Project Gutenberg\footnote{\url{https://www.gutenberg.org/}}, with two filter criteria: (1) the language was Finnish, and (2) the text was in utf-8 plain text format. There is currently no method for filtering out translated works so we estimate that the dataset consists of approximately 50\% translated texts. The data is publicly available\footnote{\url{https://github.com/esohman/FinLit-corpus}}.

We used the simple gutenberg-cleaner\footnote{\url{https://libraries.io/pypi/gutenberg-cleaner}} to get rid of the preamble and the legal text at the end of the book, then we created a regex to extract key information such as the title, the name of the author, the year of publication, and whether the book was originally written in Finnish. The translation status of the book was extracted based on whether the terms \textit{suomentaja}, \textit{suomennettu}, \textit{suomentanut}, or any version of \textit{kääntäjä/käännös/käännetty} etc.\footnote{All the terms roughly translate to "translator" or "translated".} were present within the first ten lines of text after the preamble was removed. 

Although we filtered books based on their encoding, a fairly large number of the books were not actually utf-8 encoded and had to be decoded and re-encoded. We used automatic encoding detection and tried to convert the texts to utf-8, but for some works this failed and in the end due to these encoding issues, our final corpus consists of 975 books instead of 1000. A vast majority (95+\%) were written or translated between the years 1850 and 1925 and over 90\% after 1880, with only a few instances of older texts, which means that the language used in the texts can be considered Modern Finnish \citep{forsman2011virtuaalinen}. The final data consists of 2,938,032 sentences and 41,417,116 tokens.

\subsection{The Emotion Intensity Lexicon}
We used the Finnish Emotion Intensity Lexicon (FEIL) \citep{ohman2022self} as a starting point for detecting affective terms\footnote{\url{https://github.com/Helsinki-NLP/SELF-FEIL}}. FEIL is based on the NRC emotion lexicon \citep{Mohammad13} and emotion intensity lexicon \citep{LREC18-AIL} and has been adapted for Finnish. It lists words alongside the emotions they are associated with as well as the intensity of the associated emotion as a number between 0 and 1. The emotions roughly correlate with Plutchik's wheel of emotions \citep{plutchik1980general} and contains the emotions \textit{anger, anticipation, disgust, fear, joy, sadness}, and \textit{trust}. We follow the best practices and ethical guidelines as set forth by \citet{mohammad-2022-ethics,mohammad2022best}. 

\section{Method}
\label{sec:meth}
Finnish is a fairly easy language to work with in terms of NLP. There are numerous high-quality resources that are actively maintained and new tools are constantly being developed for various written standards of Finnish and Finnish-adjacent languages. Nearly all of these resources are also open-source and freely available \citep{hamalainen-alnajjar-2021-current}. Therefore, we were privileged enough to have several different lemmatizers and tokenizers at our disposal, some even specifically made for older Finnish texts. We also added to this list by creating a Finnish version of the chapterize Python package\footnote{The original: \url{https://pypi.org/project/chapterize/} and the Finnish version: \url{https://github.com/esohman/chapterize-fi}} package.

 Once the data was cleaned, we preprocessed it by lemmatizing, splitting it into paragraphs, and tokenizing the texts. 
 We attempted to fine-tune Finnish BERT \citep{virtanen2019multilingual} to work with our texts (as per \citealt{gururangan-etal-2022-demix}), but the vocabulary was not improved sufficiently to work with data so different from the original training data for Finnish BERT. In order to find the best tool for our data we tried multiple different lemmatization tools. These tools included the Turku Neural Parser \citep{kanerva2019universal}, murre \citep{partanen-etal-2019-dialect,hamalainen-etal-2021-lemmatization}, and both the \textit{experimental} and \textit{news} Finnish spaCy models. In the end we settled for the Turku Neural Parser as the results were the most accurate (see table \ref{tab:eval} for an example) and all words were not only parsed but parsed correctly in context as well. 
 
 In the example, the Turku Neural Parser was the only one able to correctly parse the nonstandard form \textit{kahvians} -- standard form: \textit{kahviansa} -- partitive case of 3rd. pers. sing./plur. coffee.  Incidentally, in the dissertation of \citet{airio2009morphological} \textit{kahviansa} is discussed as an example of "parasite words" since it can be mistakenly split into \textit{kahvi} (coffee) and \textit{ansa} (trap), something none of the lemmatizers did. With careful optimism, we take this as a demonstration of how good lemmatizers for morphologically complex languages have become in the past decade.

\begin{table}
 \begin{adjustbox}{max width=\textwidth}
  \newcolumntype{+}{>{\global\let\currentrowstyle\relax}}
  \newcolumntype{^}{>{\currentrowstyle}}
  \newcommand{\rowstyle}[1]{\gdef\currentrowstyle{#1}%
    #1\ignorespaces
  }

  \centering
  \begin{tabular}{+>{\bfseries}ll}
    \hline
    \rowstyle{\bfseries}

Translation & The provost sits down in his rocking chair, stands his pipe on the floor against the table leg, and starts drinking his coffee \\
Original               & Rovasti istuutuu keinutuoliinsa, panee piippunsa lattialle pöydän jalkaa vasten pystyyn ja rupeaa juomaan kahvians \\
Murre (hist)           & (öljy)movasti istua keinutuoli panna pippu latija pöytä jalka vaste pystyä ja ruveta juoma kahvis                  \\
spaCy (news\_lg)       & Rovasti istuutua keinutuoliinsa  panee piippu lattia pöytä jalka vasten pystyyn ja rupeata juoda kahvians          \\
spaCy (exp. /w voikko) & rovasti istuutua keinutuoli  panna piippu lattia pöytä jalka vasten pystyyn ja ruveta juoda kahvians               \\
UralicNLP              & Rovasti|rovasti istuutua panna piippu lattia pöytä jalka vasten pystyyn|pysty ja ruveta juoma|juoda                \\
Turku Neural Parser    & rovasti istuutua keinu\#tuoli panna piippu lattia pöytä jalka vasten pystyyn ja ruveta juoda kahvi \\
Ideal lemmmatization & rovasti istuutua keinutuoli panna piippu lattia pöytä jalka vasten pystyyn ja ruveta juoda kahvi\\
\hline
\end{tabular}
\end{adjustbox}
\caption{Example of lemmatization using different lemmatizers for Finnish.}
\label{tab:eval}
\end{table}

The different lemmatizers had different strengths and weaknesses. For example, spaCy is very easy to use and install, it is reasonably fast for this amount of text, and also produces other information that most of the other lemmatizers leave out. It is therefore great for exploring other aspects of the data such as NER and similarity scores. However, the Turku Neural Parser produced the best results in an easy-to-use pipeline\footnote{\url{http://turkunlp.org/Turku-neural-parser-pipeline/}}. Omorfi \citep{pirinen2015omorfi} and FinPos \citep{silfverberg2016finnpos} might also have been good candidates for lemmatizers, but these proved difficult to install on the platforms available for this project.

Once we had the texts lemmatized and tokenized, we attempted to identify the first three paragraphs of each text. This was a more complex process than expected as even after removing the preambles/headers and footers, miscellaneous metadata of various shape cluttered the start of the book. There was very little uniformity or even commonly recurring patterns of where the actual text of the book or chapters starts. This led us to create a Finnish version of the chapterize package for Python, which enabled use to split texts into chapter and recognize opening paragraphs when used in conjunction with the sentence and paragraph ids provided by the conllu metadata generated by Turku Neural Parser. We used two different text sections as targets for overall mood detection: the first three paragraphs of each book, and the first 200 tokens from each chapter in each book. 

The size of the bins was chosen based on findings by \citet{teodorescu2022frustratingly} that suggest even at a few hundred tokens, lexicon-based methods can excel at estimating emotion arcs beyond current state-of-the-art machine learning capabilities. They suggest that lexicon-based approaches are more suitable "for applications where simple, interpretable, low-cost, and low-carbon-footprint systems are desired" \citep[8]{teodorescu2022frustratingly}.

Previous studies \citep{ohman2020litemo} have made it clear that although the results from lexicon-based emotion detection can be very accurate in terms of real-world congruency, certain words can easily obfuscate the results. One of the literary works used as quality assessment and proof-of-concept the novel \textit{Rautatie} (tr. as \textit{The Railroad}) by \textit{Juhani Aho} focuses on the novelty of a railroad track coming to a peripheral rural village and naturally contains many instances of the word \textit{railroad}, the fact that the word itself was associated strongly with \textit{trust} in the lexicon skewed the results and were not representative of the level of overall \textit{trust} in the novel. Therefore we opted to remove the term from the lexicon. 

On the other hand, FEIL contains mostly contemporary words and their contemporary emotion associations. It is reasonable to assume that these words and their associations would have been subject to semantic shift. Hence we needed to ensure that first and foremost the most common words in our texts that in our subjective opinion have an emotion association are indeed in the lexicon, and secondly that the most common words that match words in the emotion lexicon are labeled correctly for emotions at reasonable intensities. These two steps are iterative and continuous in that they should be repeated whenever the lexicon or lemmatization is altered. 

Removing words that carry little to no emotion association is fairly unproblematic, however, altering emotion labels or intensity scores or adding new words to the lexicon should be avoided. If necessary, such additive alterations require the use of multiple annotators who are not the authors and cross-checking the results using inter-annotator agreement scores \citep{van2021validity}.

For this reason, we did not want to introduce new biases into the lexicon based our own interpretations of emotion intensities. Thus we developed an alternative method that would allow us to easily and objectively add words to the lexicon. To enable this we created word embeddings of our texts and used them to look up words in the lexicon with high cosine similarity to the words we wanted to introduce to the lexicon (as per e.g. \citealt{maas-etal-2011-learning,yu2017refining,ye-etal-2018-encoding}). This lead to the association for e.g. the words \textit{kirkas, valkoinen, and valkea} (clear, white, white/light/bright often referring to fire or morning light) to be identical. The words that needed to be added were relatively few so we were able to manually check that their emotion associations and intensities made sense. For future projects, we intend to employ human annotators in conjunction with word embeddings. Nonetheless, this approach alone showed much promise and was very accurate within the small sample size in both recognizing semantically similar words and in mitigating issues with semantic shift.

Ultimately we ended up removing 128 entries from the lexicon and adding 203 tokens including \textit{rakastaa}, to love. As the FEIL lexicon was based on an English lexicon, this exlusion of such a central emotional term demonstrates the issue with noun and verb distinctions in Finnish compared to English. The noun and verb forms are often the same in English unlike in Finnish where the forms are distinct (cf. to love/a love, to run/a run vs. rakastaa/rakkaus, juosta/juoksu). Many of such instances were addressed in the creation of FEIL with the addition of verb forms copying the intensity scores and emotion associations of the English words, yet many verb forms are still missing from the lexicon. 

We used this domain- and period-specific version of FEIL to tabulate normalized (per token count for inter-text comparability), intensity scores for each target text. Other future projects should include checking that both noun and verb forms are found in the lexicon.

\begin{figure*}
    \centering
    \includegraphics[scale=.6]{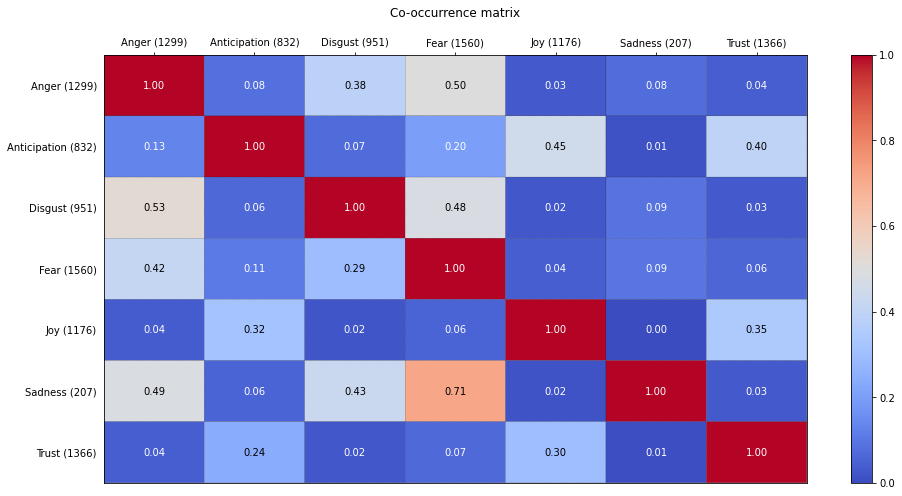}
    \caption{Co-occurence of emotions in FEIL.}
    \label{fig:corr}
\end{figure*}

The co-occurence matrix shows that emotion associations are not linear. A word associated with \textit{anger} is quite likely (0.5) to also be associated with \textit{fear}, but a word associated with \textit{fear} is slightly less likely to be associated with \textit{anger} (0.42). Words associated with \textit{sadness} are highly likely to also be associated with \textit{fear}, however, due to the very low number of \textit{sadness}-associated words in the lexicon, \textit{fear}-associated words are much less likely to simultaneously also evoke \textit{sadness} (0.09). \textit{Joy} is also the only purely positive emotion in the lexicon, which means it is more likely to be associated with many positive emotions beyond \textit{joy} itself due to annotation process conducted with best-worst scaling \citep{kiritchenko-mohammad-2017-best} (ranking words associated with a specific emotion in terms of least to most associated).

As mentioned earlier, tone and mood can be difficult to distinguish from each other and can even be intertwined to different degrees, especially if we focus on words alone. However, we argue that the tone of a literary text tends to shift much more within and between chapters, and therefore by focusing on the first paragraphs of each chapter, or even the opening paragraphs of the first chapter only, we can get a fairly accurate idea of the mood of the text due to its more stable nature, with a smaller risk of it being confused with tone.

It is well-established that first impressions matter in literature and beginnings tend to shape the experience of reading and set the mood for the whole text. Tone might vary within one text depending on changing narrative viewpoints or even narrators, or switching from descriptive language ot dialogue. Theories of perception (e.g. \citealt{perry1979literary}) argue that the openings play a crucial role in creating a text’s overall emotional disposition. The affective language and emotional effects created at the beginning of a text modify and adjust the reader’s general emotional orientation by shaping up modes of perception and organization of information. For instance, the melancholic mood created in the beginning of Aho’s \textit{Rautatie} (Railroad), or the strong effects of disgust in the beginning of Sillanpää’s \textit{Hurskas kurjuus} (Meek Heritage), are likely to influence the reader experiences later reactions and feelings triggered by narrative events.

\section{The Mood in Selected Texts}
As the results mostly consist of a dataframe with emotion intensity scores for each text, the results do not easily lend themselves to visual representations. Therefore, we are focusing the presentation of our results on a small subset of data. We chose four texts based mostly on the second author's area of expertise and previous intimate analyses of the affective landscapes of these texts. The scores for the chosen texts' first three paragraphs are presented in table form in table \ref{tab:num} and f0or both approaches comparatively in figures \ref{para3} and \ref{allch}.

\begin{table}
  \newcolumntype{+}{>{\global\let\currentrowstyle\relax}}
  \newcolumntype{^}{>{\currentrowstyle}}
  \newcommand{\rowstyle}[1]{\gdef\currentrowstyle{#1}%
    #1\ignorespaces
  }

  \centering
  \begin{tabular}{+>{\bfseries}llllllll}
    \hline
    \rowstyle{\bfseries}
\textbf{title}  & \textbf{anger} & \textbf{anticipation} & \textbf{disgust} & \textbf{fear} & \textbf{joy} & \textbf{sadness} & \textbf{trust} \\ 
Putkinotko      & 4.22           & 8.67                  & 5.42             & 11.75         & 15.69        & 1.47             & 22.07          \\
Kauppa-Lopo     & 12.45          & 5.71                  & 7.40             & 19.11         & 7.66         & 0                & 0              \\
Rautatie        & 8.56           & 21.15                 & 0                & 10.50         & 13.81        & 5.52             & 16.74          \\
Hurskas kurjuus & 20.33          & 9.90                  & 11.25            & 27.73         & 9.07         & 7.29             & 15.05        \\
\hline
\end{tabular}
\caption{Normalized emotion scores for each novel's first three paragraphs.}
\label{tab:num}
\end{table}

\subsection{An overview of the selected texts}

The first one is Juhani Aho’s breakthrough novel \textit{Rautatie} (tr. as \textit{The Railroad}, 1884). In this text, from the perspective of the implied reader, the novel evokes emotional effects of melancholia and nostalgia, which are characteristic of Aho's work.

The second one is Minna Canth’s \textit{Kauppa-Lopo} (no translation, the title refers to the protagonist's nickname, 1889), a tragic story of poverty and illness. The beginning of the novella, set in prison, underlines the anti-hero's ugly appearance, but the narrative contrasts the physical ugliness with an inner goodness: she is described as good-hearted and compassionate towards other people. The story was met with anger in contemporary audience, and the critics did not value her social criticism. Canth's naturalism was considered "poor art" by her contemporaries and she was accused of being an admirer of disgust, “destroying the laws of beauty, unfolding ugliness in every sense” \citep[52]{rossi2007}.

The third one is Frans Emil Sillanpää’s \textit{Hurskas kurjuus} (tr. as \textit{Meek Heritage}, literally “Sacred Misery”, 1919), which  begins with a shocking prologue which anticipates the death of the protagonist: it describes the execution of a poor tenant farmer who had ended up as a Red Guard soldier in the Finnish Civil War (1918). Despite the negative emotions and the tragic events the narrator also expresses trust and comfort in the future of the Finnish nation. When Sillanpää’s novel appeared it was met with confusion.

The fourth and last one is \textit{Putkinotko} (no translation, 1919-20) by Joel Lehtonen. Like Sillanpää’s \textit{Sacred Misery}, this novel tracks the tensions that escalated in the Finnish Civil War in 1918. The novel’s protagonist, a good-hearted yet self-willed tenant farmer resigns to obey the landlord and instead resorts to illegal distillery to support the family. The novel is emotionally ambivalent: the idyllic descriptions of Finnish summer nature and the comic elements are likely to arouse positive emotions, while the unembellished description of poverty intends to evoke moral anger and sadness over social inequality.

\subsection{Results}
\begin{figure}[htbp!]
\centering
\includegraphics[scale=1]{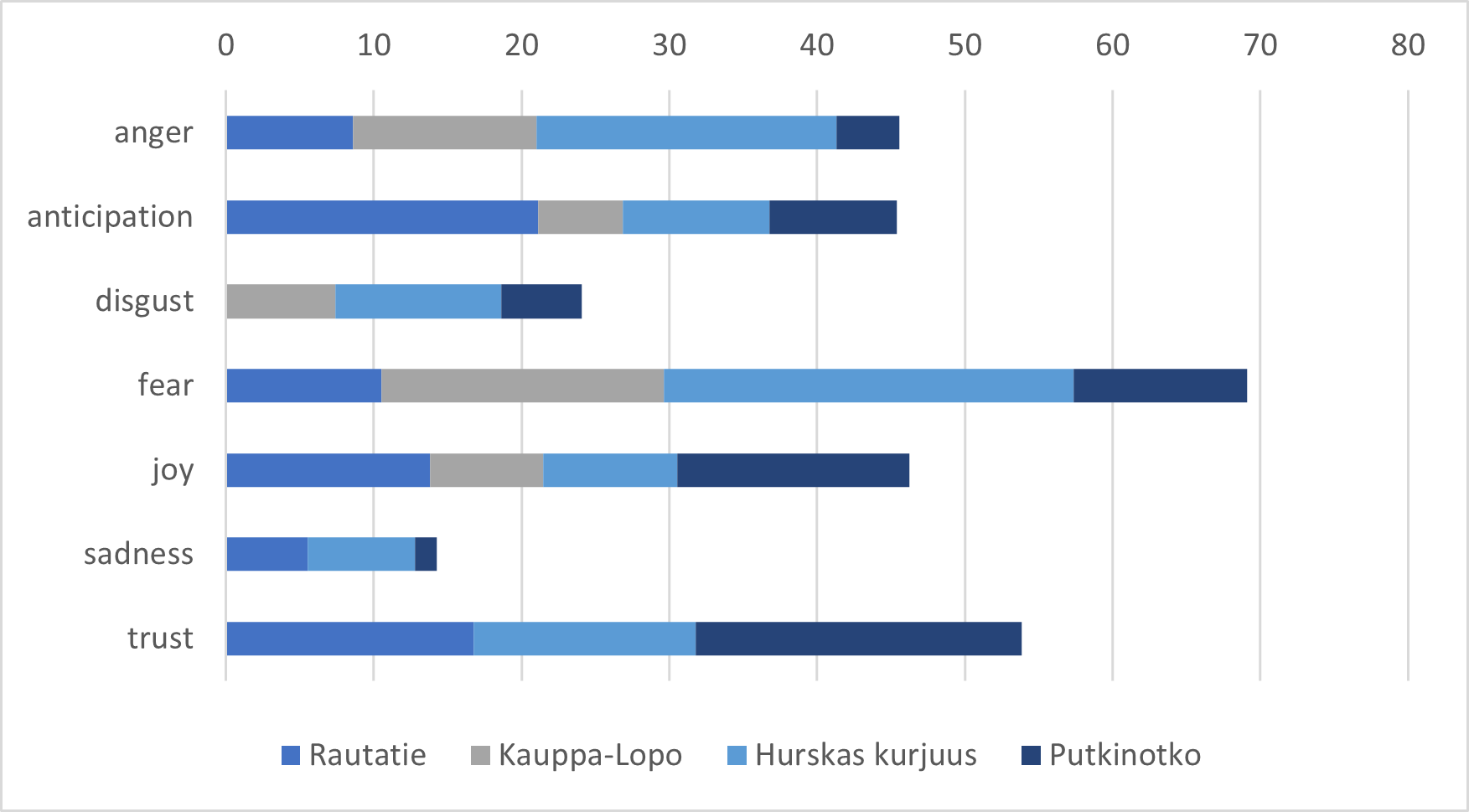}
\caption{Emotion word distribution in first three paragraphs per 1000 words}
\label{para3}
\end{figure}

\begin{figure}[htbp!]
\centering
\includegraphics[scale=1]{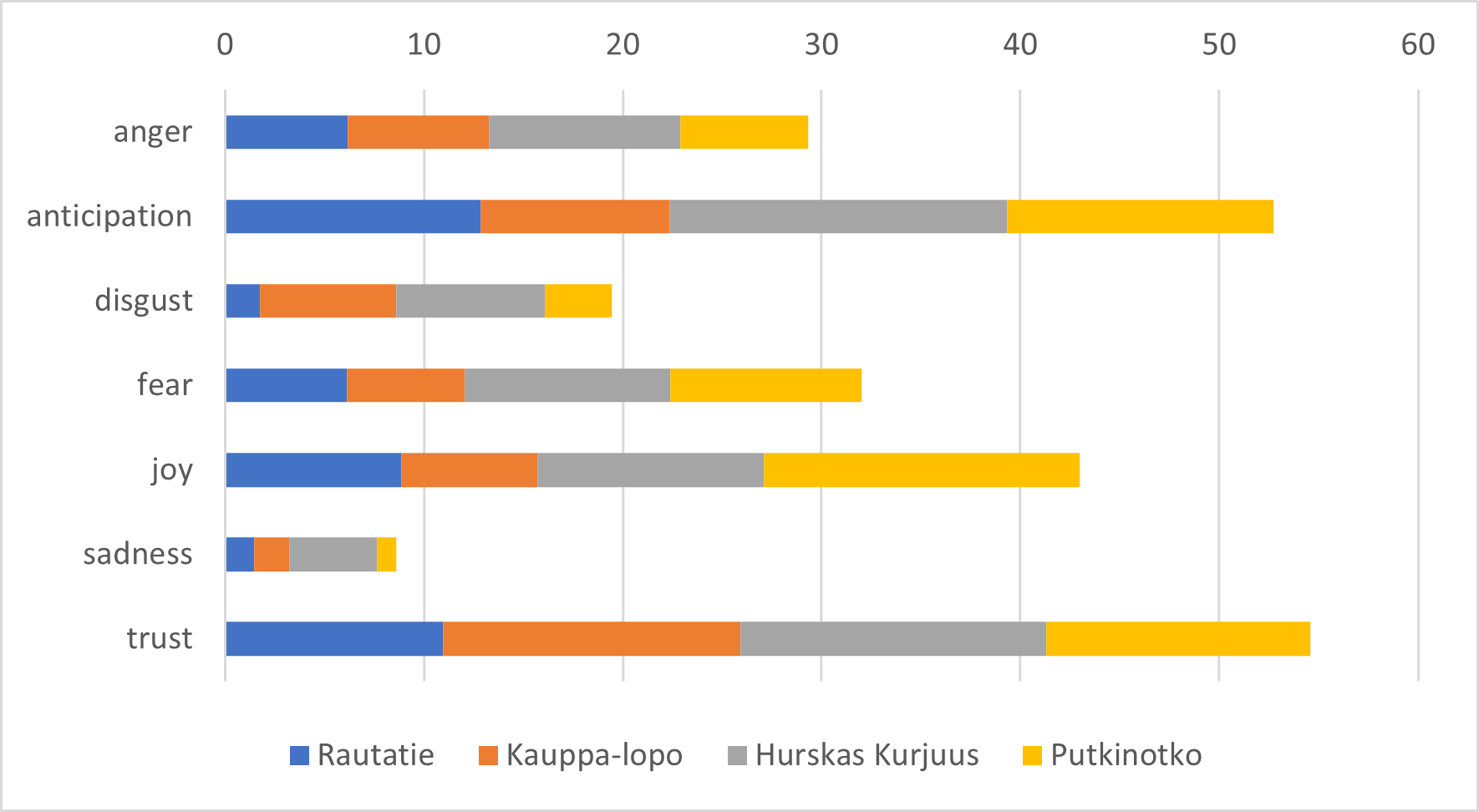}
\caption{Emotion word distribution in the first 200 tokens of each chapter per 1000 words}
\label{allch}
\end{figure}

The results for the texts are presented in figures \ref{para3} and \ref{allch}. 

 In figure \ref{para3} we can see some patterns emerge. In particular, the laconism of \textit{Rautatie} is evident when compared to the other authors and the strong early emotional impact of \textit{Hurskas Kurjuus} becomes very apparent, with \textit{fear} and \textit{anger}, but also \textit{sadness} being particularly notable. \textit{Fear} and to some extent \textit{anger} are also highly present in \textit{Kauppa-Lopo}, likely due to the descriptions of the prison environment and the appearance of the protagonist. \textit{Trust} and \textit{joy} are the most notable emotions in \textit{Putkinotko}, perhaps due to the detailed descriptions of the idyllic landscape that dominate the opening chapter.

 When comparing the two different segmentations, first paragraph-only vs. opening paragraphs of each chapter, the latter shows a stronger prevalence of positive emotions. We take this to indicate that the intended opening mood for these novels is constructed on negative emotions intended to evoke strong feelings in the reader. In the former segmentation, the emotion distribution is more varied and seemingly converging on the distribution of emotion words in the lexicon.

\section{Discussion}
Our computationally derived results correspond highly with qualitative evaluations of the same target texts in terms of established interpretations of mood specifically when compared to the emotion word distribution of the first three paragraphs of a literary text. The valency and intensity of emotions in the opening paragraphs of the book and the opening paragraphs of all chapters the emotions approach the distribution of emotions in the lexicon in the former segmentation. That is, they become muddled because they are not focused enough. Only minor effects of the authors' style can be discerned when applying this method across all chapters. Not only are the differences between chapters reduced with the former, all-chapters, approach, but the differences between the texts also become less clear. This could also be in part because the focus and narrative choices become more varied and therefore the results average out and start to converge on the distribution of emotions in the lexicon. We recommend that the quest for mood should begin with the opening paragraphs of a text.

The qualitative analysis of the selected works, summarized above, has paid attention to various aspects of depicting and triggering emotions in a literary texts: 
\begin{enumerate}
     \item the characters' and the narrators' emotions
     \item  the emotional effects targeted at the implied reader
     \item the empirical readers' reactions in contemporary reception 
     \item the texts' tone, the organizing feeling of a literary work, which is never reducible to a reader's emotional response to a text, nor to a text's internal representations of feeling (on the concept of the emotional effect, see \citealt{lyyti2017written}; on the notion of tone, see \citealt[28]{ngai2005ugly}. 
 \end{enumerate}
 
It should be noted that evoking emotional effects in literature is not restricted to emotion-associated words or to direct descriptions of the character's emotions. All facets of the narrative, from the description of objects to the narrative point of view and style, including tropes and even the rhythm of the text are important aspects in triggering emotional effects in the reader. As an example, the melancholic tone of Juhani Aho's text is not generated by themes of separation and loss alone but also by Aho's style, which favors fragmentation and loosening of syntax, with a recurring mannerism of three points "...",  as a sign of hesitation and withdrawal, even evoking a depressive loss of contact. 

The qualitative analysis demonstrates that the selected texts depict and trigger negative emotions in particular: feelings of deception, fear, anxiety, disgust and hatred, anger, moral indignation and melancholia. This can be partly explained by genre-specific emotional effects: a critical naturalist novel tends to shock and challenge its reader by representing and inciting strong negative emotions, which confirm the effect of the reality of a text and direct the reader's attention to the social defects described. For instance, the emotion of \textit{disgust}, a genre-specific emotion of the naturalist novel, is a named emotion and salient in Sillanpää's and Canth's novels in particular \citep{rossi2007,rossi2017writing,rossi2020}.  

The salience of negative emotions can be explained by the importance of negative emotions in literature and art in general. As discussed by \citet{menninghaus2017distancing} negative emotions are an important resource for the arts, since negative emotions have been shown to be particularly powerful in securing attention, intense emotional involvement, and high memorability, and hence is precisely what artworks strive for. Narrative plots routinely involve social conflicts and both represent and elicit negative emotions in response to such conflicts: failing marriages, unhappy love, long separations, adultery, betrayed friendship, and the like. In narratives, happiness is generally not described in great detail but rather evoked as a peak moment to be challenged, or as a goal to strive for: Canth and Jotuni depict the characters' desire for happiness and love, which is not possible in today's society; Sillanpää's narrator expresses trust in and hope for the future of Finland, and Aho concludes \textit{The Railroad} with the normative happy ending of a fairy tale - yet this happiness is only evoked, but not described in detail.

\section{Future Work}
This dataset will be used for more robust detection of tone and mood in Finnish literature using many novel approaches. The approaches developed for this dataset will also be used with other literary datasets in many other languages. Our preliminary studies show that the "big data" results support qualitative analyses and further justify the use of purely lexicon-based methods and affect as a proxy for mood when dealing with larger collections of text where word choice is an important factor in evoking affective states in the reader. Explicitly, we can see that the choice of emotion-associated words in the first three paragraphs correlates highly with established analyses of mood in the selected texts. We hope to add established emotion categories from literary affect studies (see e.g. \citealt{hogan2011literature}) to the lexicon as a measure to further improve the usability of the FEIL lexicon for the literary domain \citep{ohman2020emotion}. Additionally, we would like to expand on the methodologies used in this exploratory study and hopefully create more and more robust approaches to tone and mood detection in literature and perhaps also incorporate intentionality detection \citep{guo2009algorithm} to further separate tone from mood.

\section*{Acknowledgements}
This work was supported by JSPS KAKENHI Grant Number 22K18154.

\bibliography{anthology,custom,jdmdh-example}
\bibliographystyle{plainnat}

\end{document}